# Automatic Extraction of Relevant Road Infrastructure using Connected vehicle data and Deep Learning Model

**Adu-Gyamfi Kojo**
Department of Civil and Environmental Engineering
Institution for Transportation, Ames, Iowa, 50010
Email: kgyamfi@iastate.edu

**Kandiboina Raghupathi**
Department of Civil and Environmental Engineering
Institution for Transportation, Ames, Iowa, 50010
Email: raghukan@iastate.edu

**Ravichandra-Mouli Varsha**
Research Scientist, Intrans
Institution for Transportation,
Ames, Iowa, 50010
Email: varsha@iastate.edu

**Knickerbocker Skylar**
Research Scientist III
Institute For Transportation
Ames, Iowa, 50010
Email: sknick@iastate.edu

**Hans Zachary N**
Research Scientist IV
Institute For Transportation
Ames, Iowa, 50010
Email: zhans@iastate.edu

**Hawkins, Neal R**
Director Research Administration
Institute For Transportation
Ames, Iowa, 50010
Email: hawkins@iastate.edu

**Sharma Anuj**
Professor Civil Construction & Environmental Eng
Institute For Transportation
Ames, Iowa, 50010
Email: anujs@iastate.edu

Word Count: 5450 words + 3 table (34 words per table) = 5,552 words

*Submitted [08/01/2023]*

## ABSTRACT

In today's rapidly evolving urban landscapes, efficient and accurate mapping of road infrastructure is critical for optimizing transportation systems, enhancing road safety, and improving the overall mobility experience for drivers and commuters. Yet, a formidable bottleneck obstructs progress - the laborious and time-intensive manual identification of intersections. Simply considering the shear number of intersections that need to be identified, and the labor hours required per intersection, the need for an automated solution becomes undeniable. To address this challenge, we propose a novel approach that leverages connected vehicle data and cutting-edge deep learning techniques. By employing geohashing to segment vehicle trajectories and then generating image representations of road segments, we utilize the YOLOv5 (You Only Look Once version 5) algorithm for accurate classification of both straight road segments and intersections. Experimental results demonstrate an impressive overall classification accuracy of 95%, with straight roads achieving a remarkable 97% F1 score and intersections reaching a 90% F1 score. This approach not only saves time and resources but also enables more frequent updates and a comprehensive understanding of the road network. Our research showcases the potential impact on traffic management, urban planning, and autonomous vehicle navigation systems. The fusion of connected vehicle data and deep learning models holds promise for a transformative shift in road infrastructure mapping, propelling us towards a smarter, safer, and more connected transportation ecosystem.

**INTRODUCTION**

Efficient and accurate mapping of road infrastructure is pivotal in optimizing transportation systems, bolstering road safety, and improving overall mobility experiences for drivers and commuters (1). However, a significant obstacle hampers progress – the time-consuming, labor-intensive process of manually identifying intersections (2). This inefficiency hinders the creation of a comprehensive and up-to-date database for the entire US road network, limiting smarter, data-driven transportation management.

Presently, intersection identification relies on visual inspection, utilizing roadway and aerial images (3), (4) based on the Model Inventory of Roadway Elements – MIRE 2.0 (5). However, this approach, as demonstrated in a recent project by Iowa DOT and the Institute of Transportation, can be resource-intensive and time-consuming. In that project, each intersection's characteristics were collected manually, showing the scale of effort required. In the aforementioned project, intersection data was initially derived from GIS-based road centerline data, with a significant portion of time devoted to manually collecting characteristics of both the intersections and their approaches. This process, while painstakingly meticulous, occasionally led to the discovery of previously unidentified intersections and the removal of non-intersection locations. These results underscore the laborious nature of the current method, the challenges in its application, and the urgency for a more streamlined and automated solution.

The Federal Highway Administration mandates that all state DOTs maintain an All-Road Network of Linear Referenced Data (ARNOLD) (*21*), which many states have utilized to derive preliminary sets of possible intersection locations. Nevertheless, despite significant efforts, maintaining ARNOLD is a continuous task. Furthermore, identifying access points and driveways, intersections with non-public roads, and network changes over time, particularly along the local system which is often challenging for state DOTs.

While this data collection methods has yielded valuable insights (6), it has proven to be resource-intensive and time-consuming, as evidenced by the 110,765 intersections gathered within a 20-month period in Iowa state alone. Although this number represents a significant effort, it falls short of covering the entirety of the vast road network. This paper proposes a novel approach based on the utilization of connected vehicle data and a deep learning model. By harnessing the power of advanced data analytics and machine learning, our proposed method aims to detect intersections and straight roads without relying on all the data details, significantly reducing the time and resources required for comprehensive road infrastructure mapping. These "straight roads" constitute the mainline portions of the road network that are not categorized as intersections. This serves as a stepping stone towards automating the entire process of generating MIRE elements.

This paper showcases the potential of our approach to accurately identify road infrastructure which include intersection and straight road segment locations automatically, even in the absence of comprehensive data, and detect possible network changes over time. Such an approach is particularly beneficial along the local system, an area often challenging for state DOTs to systematically identify, thereby emphasizing the feasibility and effectiveness of our proposed method for automatic road infrastructure extraction. By leveraging the rich spatial and temporal information present in driver trajectories, we overcome the limitations of traditional data collection methods, providing a scalable solution for road network analysis. The utilization of connected vehicle data, which is available US-wide and can be extended globally, combined with cutting-edge deep learning models, enables us to accurately identify different road segments from vehicle trajectories. This scalable approach not only sets the stage for potential impact across the US but also globally.

The authors envision a future where the integration of connected vehicle data and deep learning algorithms will revolutionize road infrastructure mapping, providing a more efficient, current, and data-driven transportation management system. This paper presents a transformative approach that sets new standards for road infrastructure mapping, paving the way for a smarter, safer, and more connected world.

**LITERATURE REVIEW**

The analysis of roadway infrastructure and the detection of intersections are fundamental data elements for various applications, including traffic management, autonomous vehicle navigation, and comprehensive road network safety analysis (*7*). Over the years, a multitude of studies have been conducted to tackle the intricate challenges inherent in this domain, employing diverse methodologies such as vehicle trajectory data analysis, GPS traces, and cutting-edge algorithms. In this literature review, we present a comprehensive overview of the significant contributions made by relevant research in this field, which mainly relies on two broad approaches - satellite imagery and vehicle trajectory features.

The use of satellite imagery for road infrastructure mapping has been prevalent for some time. Researchers have employed remote sensing images to perform automatic or semi-automatic road network extraction (*8*)(*9*)(*10*). These methods usually rely on various image processing techniques, including edge detection, morphological operations, and region growing, among others (*10*)(*11*). More advanced techniques, such as convolutional neural networks, have also been applied to extract multiple objects from aerial imagery (*12*). However, the main limitations of these approaches include the difficulty in distinguishing roads from similar-looking features in satellite images, such as rooftops and parking lots (*9*). Furthermore, the high cost of acquiring high-resolution satellite imagery and the inability to provide real-time updates on changes in road infrastructure are other significant challenges associated with these methods (*13*).

A different approach to road infrastructure mapping focuses on extracting relevant features from vehicle trajectories. This method leverages GPS data collected from vehicles to create or update road maps (*14*)(*15*)(*16*). For example, Saldivar-Carranza and Bullock's 2023 study presented a data-driven method for mapping intersection geometry to enhance the scalability of trajectory-based traffic signal performance measures (*15*). They used high-resolution GPS data and applied a series of algorithms to detect and classify intersection elements, which were then used to create intersection maps. Their work highlights the potential of vehicle trajectory data not only in road infrastructure mapping but also in optimizing traffic signal performance. However, their method still relies on predefined intersection templates, which could limit its adaptability to intersections with unique or complex designs. Moreover, their study did not address how to distinguish between different types of road infrastructures beyond intersections.

Vehicle trajectory-based methods have the advantage of providing real-time updates and the ability to map areas that are not visible in satellite images due to building occlusion or dense tree canopy. However, the accuracy of these methods depends largely on the quality and quantity of the collected GPS data, which is typically noisy and sparse. Therefore, sophisticated algorithms are required to clean and interpolate the data before road features can be accurately extracted (*15*). Moreover, these approaches often struggle with complex road structures.

The proposed methodology in our paper builds upon the aforementioned methods, incorporating the strengths of each while mitigating their drawbacks. It leverages connected vehicle data, which provides extensive, real-time, and large-scale GPS data, overcoming the limitations of sparsity and inaccuracy (*17*). In addition, our method employs deep learning models, specifically the YOLOv5 algorithm, which has demonstrated excellent performance in image classification tasks, to accurately classify road segments and intersections (*18*).

Instead of extracting geometric features from vehicle trajectories or performing image processing on satellite images, our approach transforms the problem into an image classification task. We segment vehicle trajectories using geohashing and generate image representations of road segments. These images are then fed into the YOLOv5 algorithm for classification. This novel approach simplifies the problem of road infrastructure extraction and enhances its efficiency, accuracy, and scalability.

The major contributions of our approach lie in its ability to automatically extract relevant road infrastructure from large-scale vehicle trajectory data with high accuracy, its scalability across the globe, and its potential for real-time updates. This is achieved by combining the strengths of connected vehicle data and deep learning models, which have not been comprehensively exploited in previous studies. The implications of our research extend to various aspects of transportation systems, such as traffic management, urban planning, and autonomous vehicle navigation.

## METHODS

The methodology section of this paper provides a detailed account of the process employed to extract road infrastructure information from connected vehicle trajectory data using geohashing (*19*) and image classification with YOLOv5 (*20*). The proposed methodology proceeds with data description and processing, the implementation of geohashing techniques to partition the trajectories onto a plot. This partitioning generates image representations of road segments. Subsequently, these images undergo processing with YOLOv5, a cutting-edge object detection algorithm, to classify and identify straight roads and intersections. The section will outline the step-by-step procedure followed, including data preprocessing, geohashing implementation, image generation, YOLOv5 configuration, and model training. Thorough explanations and justifications will be provided to validate the effectiveness and accuracy of the proposed approach. The ultimate objective of this research is to improve the accuracy and efficiency of mapping roadway infracstructure to then advance agency capabilities in terms of network safety analysis, traffic management, and autonomous vehicle navigation systems.

### Data Description

In the present research, we leveraged the Wejo connected vehicle dataset, a leading data source in the transportation industry, to extract valuable road infrastructure information from vehicle trajectories specifically in the Ames area within Story County, Iowa (**Figure 1**). The dataset for this study was chosen to encompass a one-day subset, providing real-time and historical data from a wide variety of vehicles, including data on GPS location, speed, acceleration, and other relevant attributes. The dataset is anonymized and aggregated, featuring vehicle trajectory data collected over a specific timeframe.

This location was selected due to its diverse road network that encompasses urban, suburban, and rural environments, thus enabling us to conduct a comprehensive analysis of road infrastructure extraction techniques.

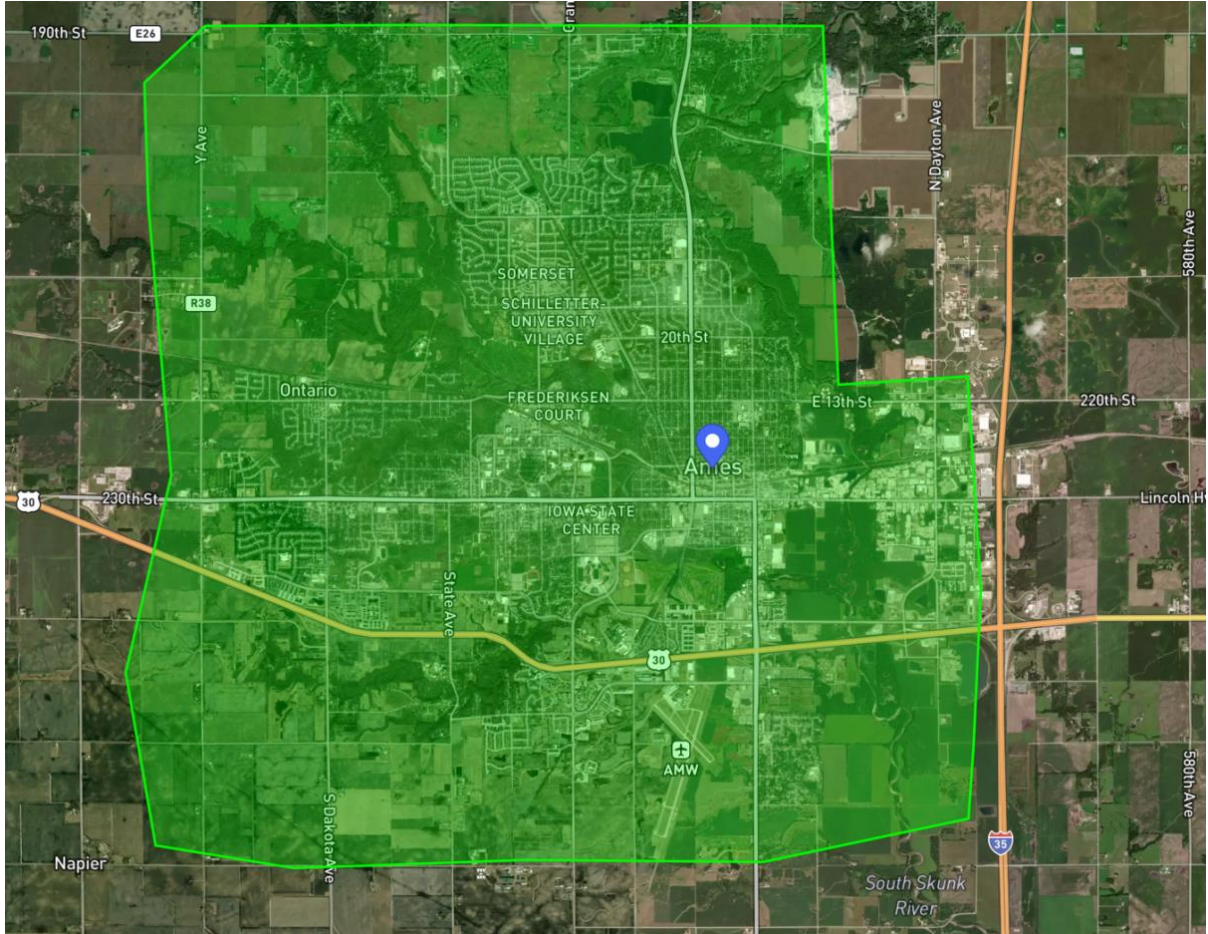

**Figure 1 Geographical area of Ames**

**Data Preprocessing**

In data preprocessing, the initial step involves removing null vehicle coordinates to ensure the integrity and accuracy of the dataset. Null coordinates can arise due to various reasons such as GPS signal loss or data transmission errors. By eliminating these null values, we aim to ensure the reliability of the subsequent analysis.

Next, we filter out the trajectories that do not follow linear roadways. Vehicles traveling off-road or on non-linear pathways can lead to inaccuracies in the representation of road infrastructure. Therefore, we cross-referenced the GPS coordinates of vehicle trajectories with Iowa DOT's Road database to identify and remove non-matching linear paths in the database. This approach, albeit relying on a fixed dataset, effectively eliminates trajectories not adhering to the roadways, and enhances the accuracy of our model. However, this step in the process can also be performed using a data-driven approach like clustering in the future. Clustering algorithms can group the data based on spatial proximity and patterns, thereby separating linear roadway trajectories from non-linear or off-road paths.

The next crucial step is to convert the vehicle data, which is initially provided as geographic coordinates, into line trajectories as shown in **Figure 2**. This conversion is necessary to represent the vehicle movements as continuous paths. To achieve this, the coordinates belonging to each unique vehicle journey ID are joined by a line, effectively connecting the consecutive points, and forming a coherent trajectory. This conversion from individual coordinates to line trajectories enables a more comprehensive representation of the vehicles' movements, facilitating subsequent analysis and visualization of the road infrastructure.

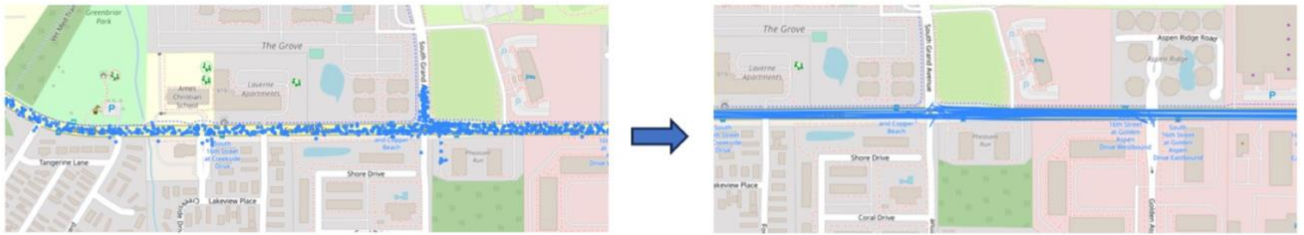

**Figure 2 Conversion of Cordinates to Trajectories**

**Geohashing Implementation**

The implementation of geohashing (*19*) techniques (**Figure 3**), which play a crucial role in the extraction of road infrastructure information from vehicle trajectories. Geohashing (*19*) is a method used to encode geographic coordinates into a string of characters, allowing for the efficient spatial indexing and retrieval of data. It provides a systematic way to divide the Earth's surface into grids of various sizes, enabling spatial operations and analysis at different levels of precision.

Geohashing employs a hierarchical structure where each level of precision corresponds to a different grid size as shown in **TABLE 1**. Geohash ranges from 1 to 12, where the lower levels grid cells cover larger areas, providing a more generalized representation, and the higher levels have granularity by dividing the surface into smaller cells. The choice of geohash precision depends on the specific requirements of the analysis, striking a balance between accuracy and computational efficiency. For the implementation of geohashing techniques in this study, a precision level of 8 was used to generate geohash codes for each vehicle coordinate.

**TABLE 1 Metric dimensions for geohash precision levels**

| Geohash Precision Levels | Grid Area (width X height) |
|---|---|
| 1 | 5,009.4km x 4,992.6km |
| 2 | 1,252.3km x 624.1km |
| 3 | 156.5km x 156km |
| 4 | 39.1km x 19.5km |
| 5 | 4.9km x 4.9km |
| 6 | 1.2km x 609.4m |
| 7 | 152.9m x 152.4m |
| 8 | 38.2m x 19m |
| 9 | 4.8m x 4.8m |
| 10 | 1.2m x 59.5cm |
| 11 | 14.9cm x 14.9cm |
| 12 | 3.7cm x 1.9cm |

Once the geohash codes are generated from the geohash level, the unique geohash codes are utilized to generate bounding boxes. These bounding boxes serve as spatial containers that encompass the corresponding trajectory segments. By defining the boundaries of each geohash cell, the bounding boxes enable the subsequent clipping of trajectories, isolating specific road segments for further analysis. The utilization of geohash codes and the generation of bounding boxes allow for the efficient organization and partitioning of the trajectory data. This approach simplifies the subsequent steps in the methodology, as it provides a structured representation of the road network, allowing for targeted analysis and processing of specific road segments. This geohash drawing on the trajectories serves as a foundational step for subsequent image generation and road segment classification.

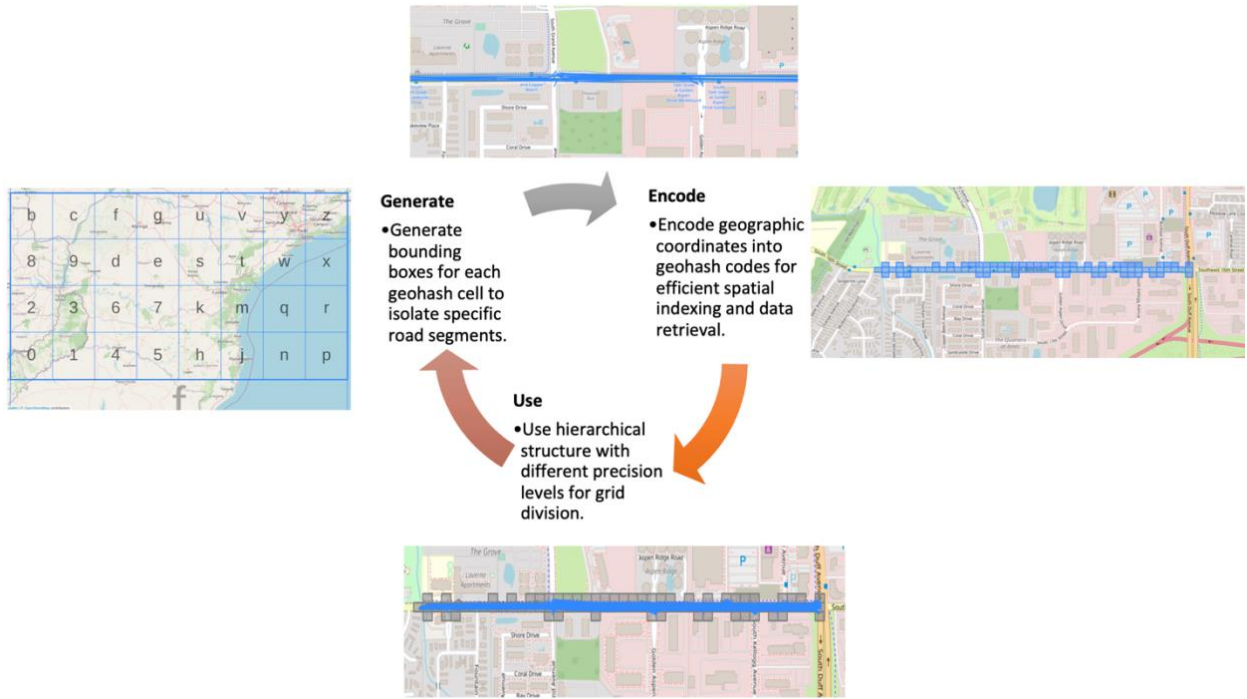

**Figure 3 Geohash Implementation Flow**

**Image Generation**

The next sub-section of the methodology generates images using the geohash boxes derived from specific geohash level (level 8). the as shown in **Figure 4**. To generate these images, the geohash boxes are employed as a means of partitioning the trajectories and extracting relevant road segments. By utilizing the bounding boxes associated with each unique geohash code, the trajectories are clipped within their respective spatial boundaries.

Once the trajectories are clipped, they are plotted on a plot figure. This plot figure serves as a visual representation of the road segments contained within each geohash box. After plotting the trajectories on the plot figure, it is saved as an image file. This image file becomes the input for the subsequent classification model based on YOLOv5. The saved images capture the extracted road segments, providing a standardized format that can be easily processed and analyzed by the object detection algorithm.

The generation of images using the geohash boxes facilitates the integration of spatial information into the image-based classification approach. By partitioning the trajectories and creating visual representations, this methodology ensures that the subsequent classification model can effectively identify and classify road segments within the images, contributing to the overall objective of extracting road infrastructure information from the vehicle trajectories.

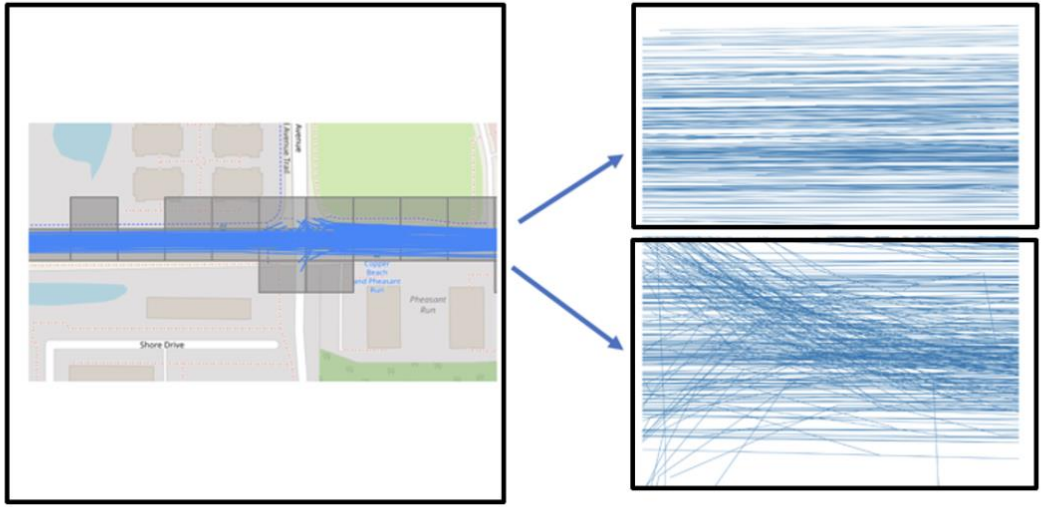

**Figure 4 Clipping trajectories for images (left) with a detailed view (right)**

**YOLOv5 Configuration**

Building upon the image generation process from the previous subtopic, the subsequent step in the methodology involves classification using the YOLOv5 model. YOLOv5 is a highly popular object detection algorithm known for its speed and accuracy. It operates by dividing images into a grid system, with each cell in the grid responsible for detecting objects within itself.

To identify intersections and straight roads from the transformed trajectory images, we utilized YOLOv5, a state-of-the-art object detection algorithm. It is important to note that the focus of this paper is not on the inner workings of YOLOv5; however, a brief overview of its role in our methodology is beneficial for understanding the overall process.

YOLOv5 is an algorithm known for its superior speed and accuracy in detecting objects within images. Unlike traditional methods, which might scan an image multiple times at different scales and locations to identify objects, YOLOv5, as the name suggests, takes only one look at the whole image. This enables it to predict multiple bounding boxes and class probabilities directly from full images in one evaluation, thus making it significantly faster and more efficient.

In the context of our work, the "objects" YOLOv5 is trained to identify are the intersections and straight roads. We trained the YOLOv5 model using labeled images of intersections and straight roads which were labeled automatically using pre existing road way intersection locations from Iowa DOT's roadway database, and once trained, the model could accurately classify these features in new, unseen trajectory images.

Although YOLOv5's application in our methodology is straightforward, its contribution to the automated extraction of relevant road infrastructure information from connected vehicle data is significant, helping us achieve high accuracy in an efficient manner.

**Model Training**

Model training is a crucial step in developing a YOLOv5 classification model for road infrastructure extraction. It involves training the model using a carefully curated dataset to learn the patterns and features associated with different road segments, such as straight roads and intersections. The following preprocessing and augmentation techniques were applied to improve the model's performance and generalization capabilities:

- Test-Train Split: The initial dataset consists of 2,217 images, which are categorized into straight roads and intersections. A test-train split is performed to ensure an unbiased evaluation of the model's performance.
- Data Preprocessing: The data are standardized through auto-orientation, resizing to a fixed size of 640x640 pixels, and converted to grayscale to reduce computational complexity.
- Data Augmentation: To increase the diversity and variability of the training data, augmentations such as flip, rotation, shear, blur, and noise were applied.

After applying these preprocessing and augmentation techniques, the dataset expanded to 5,900 images. This augmented dataset provided a rich training set with a variety of road infrastructure instances, enabling the model to learn and generalize better.

Through these preprocessing and augmentation techniques, the YOLOv5 classification model was prepared for training. The considerations of test-train split, data preprocessing, and data augmentation collectively contribute to improving the model's accuracy, robustness, and ability to handle real-world road infrastructure extraction scenarios.

In conclusion, the methodology presented in this paper encompasses a systematic approach to extract road infrastructure information from connected vehicle trajectory data. The combination of geohashing techniques and image classification with YOLOv5 provides a comprehensive framework for analyzing and identifying road segments, including straight roads and intersections. The step-by-step procedure outlined, from data preprocessing to geohashing implementation, image generation, YOLOv5 configuration, and model training, ensures a thorough and effective process for enhancing road network analysis, traffic management, and autonomous vehicle navigation systems. To visualize the overall steps of the methodology, a flow diagram as shown in **Figure 5** is provided, offering a clear representation of the sequential process. The subsequent section will present the results and discussion, showcasing the effectiveness and accuracy of the proposed methodology in extracting road infrastructure information from connected vehicle trajectory data.

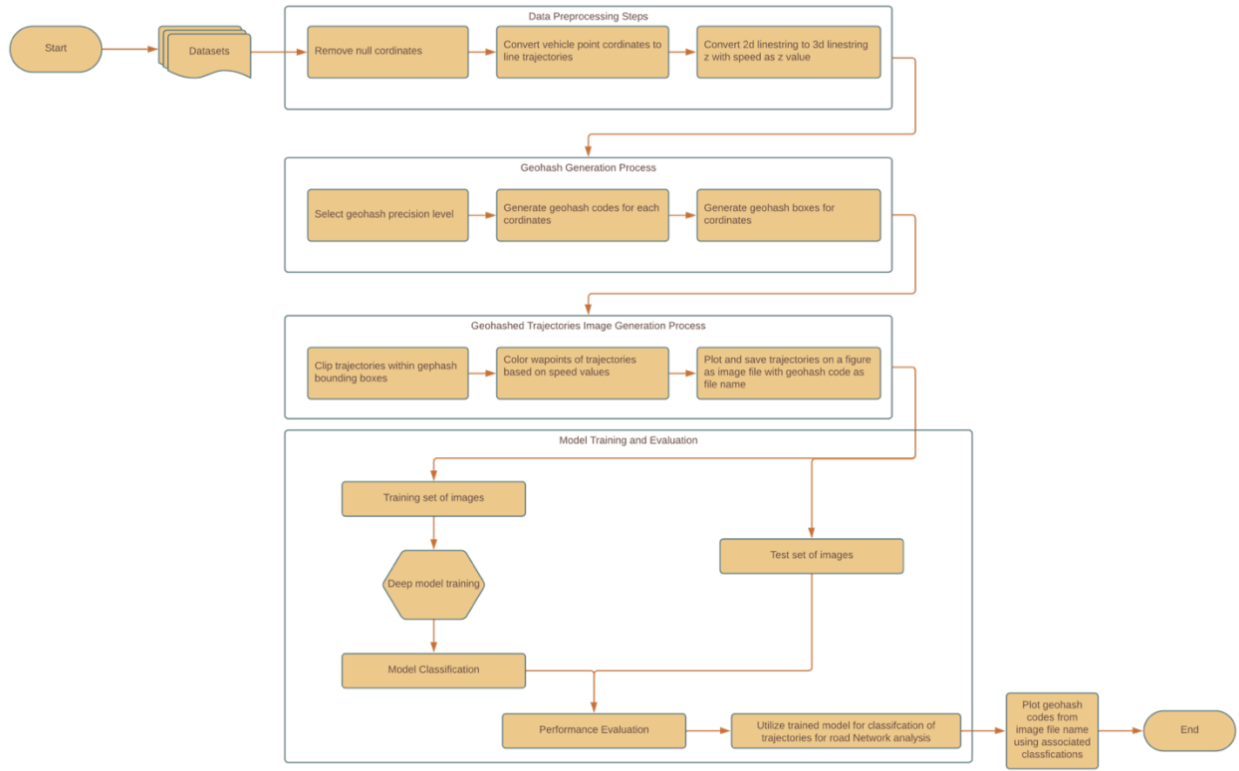

**Figure 5 Flow Chart of the Methodology**

**RESULTS**

This section presents and analyzes the results of the study, focusing on the effectiveness, accuracy, and potential implications of the proposed approach. The performance of the YOLOv5 classification model, trained using the methodology, is evaluated, and the findings are discussed in detail.

**Performance Metrics**

The initial results of the YOLOv5 classification model, trained on the dataset without colored images, are presented in **Table 2**. The table provides performance metrics for two classes of road infrastructure: "Intersection" and "Straight." The metrics evaluated include precision, recall, F1-score, and support.

**TABLE 2 Performance metrics for the model without colored images.**

|  | *Precision* | *Recall* | *F1-score* | *Support* |
|---|---|---|---|---|
| Intersection | 0.73 | 0.69 | 0.71 | 55 |
| Straight | 0.90 | 0.92 | 0.91 | 166 |
| Accuracy |  |  | 0.86 | 221 |
| Macro Avg | 0.82 | 0.80 | 0.81 | 221 |
| Weighted Avg | 0.86 | 0.86 | 0.86 | 221 |

For the "Intersection" class, the model achieved a precision of 0.73, indicating that 73% of the predicted instances classified as intersections were accurate. The recall for this class was 0.69, meaning the model correctly identified 69% of the actual intersections. The F1-score, which combines precision and recall, was calculated as 0.71. The support column indicates that there were 55 instances of the "Intersection" class in the dataset.

Regarding the "Straight" class, the model achieved a precision of 0.90, suggesting a high level of accuracy in identifying straight road segments. The recall for this class was 0.92, indicating that the model successfully detected 92% of the actual straight road segments. The F1-score for the "Straight" class was computed as 0.91. The support column shows that there were 166 instances of straight road segments in the dataset.

The overall accuracy of the model, calculated as 0.86, indicates that the model correctly classified 86% of the road infrastructure instances in the dataset.

**Model Performance with Colored Images**

To further improve the model's performance, a new dataset consisting of images with color-coded trajectories based on the waypoint speed of vehicles within each geohash was used for retraining the YOLOv5 model. The performance metrics for the model trained on these colored images are presented in **Table 3**.

**TABLE 3 Performance metrics for the model with colored images.**

|  | *Precision* | *Recall* | *F1-score* | *Support* |
|---|---|---|---|---|
| Intersection | 0.98 | 0.84 | 0.90 | 67 |
| Straight | 0.94 | 0.99 | 0.97 | 170 |
| Accuracy |  |  | 0.95 | 237 |
| Macro Avg | 0.96 | 0.91 | 0.93 | 237 |
| Weighted Avg | 0.95 | 0.95 | 0.93 | 237 |

For the "Intersection" class, the model achieved a significantly higher precision of 0.98, compared to the previous precision of 0.73 without colored images. This indicates a substantial improvement in accurately identifying intersections. The recall for this class also increased to 0.84, showing that the model detected a greater portion of actual intersections compared to the previous recall of 0.69. The F1-score for intersections was calculated as 0.90, demonstrating a notable improvement from the previous F1-score of 0.71. The support column shows that there were 67 instances of intersections in the dataset.

Regarding the "Straight" class, the precision value increased to 0.94, showcasing a remarkable improvement compared to the previous precision of 0.90 without colored images. This indicates the model's enhanced accuracy in identifying straight road segments when trained on the new dataset. The recall for this class significantly improved to 0.99, surpassing the previous recall of 0.92, and indicating that the model successfully detected nearly all of the actual straight road segments. The F1-score for the "Straight" class was computed as 0.97, which is a notable enhancement compared to the previous F1-score of 0.91. The support column indicates that there were 170 instances of straight road segments in the dataset.

The overall accuracy of the model trained on the new dataset was 0.95, a substantial improvement compared to the previous accuracy of 0.86 without colored images. This higher accuracy suggests that the incorporation of average speed information within geohashes has significantly contributed to the model's ability to classify road infrastructure instances accurately.

**Confusion Matrix Analysis**

To evaluate the model's performance on unseen data, confusion matrices were generated for both the model without colored images and the model with colored images. The confusion matrix for the model without colored images is shown in **Figure 6**, while the confusion matrix for the model with colored images is shown in **Figure 7**.

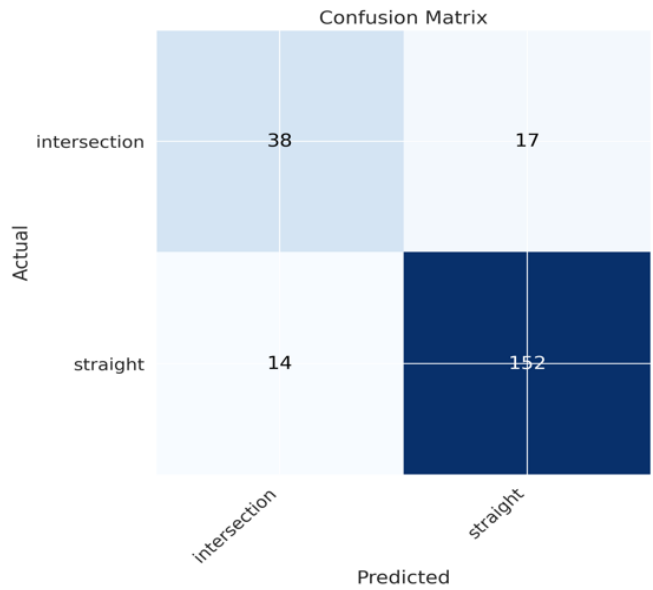

**Figure 6 Confusion matrix for the model without colored images.**

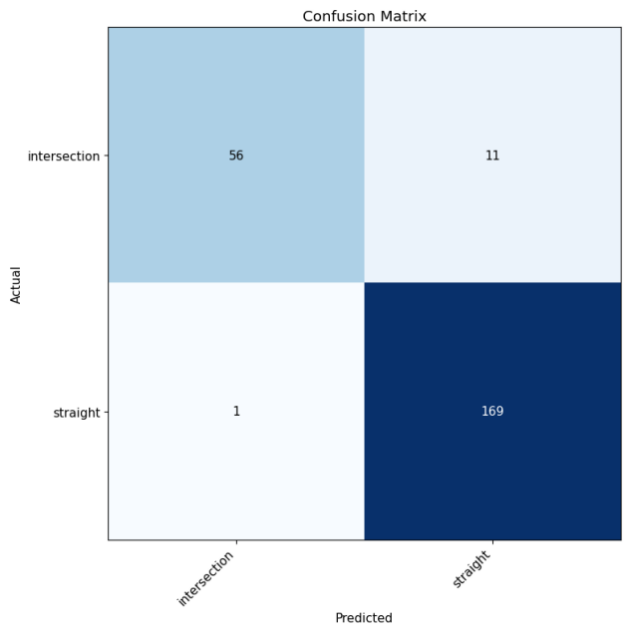

**Figure 7 Confusion matrix for the model with colored images.**

Comparing the two confusion matrices, it is evident that the model trained with colored images and speed information achieves better classification performance. The number of true positives for both classes has increased, with 56 instances of "Intersection" correctly classified and 169 instances of "Straight" correctly classified. The number of false negatives has decreased, indicating a better ability to detect instances of both classes. The overall performance improvement is apparent from the updated confusion matrix. A sample of satellite imagery of detected intersections and straight roadway segments are shown in the **Figure 8** and **Figure 9**.

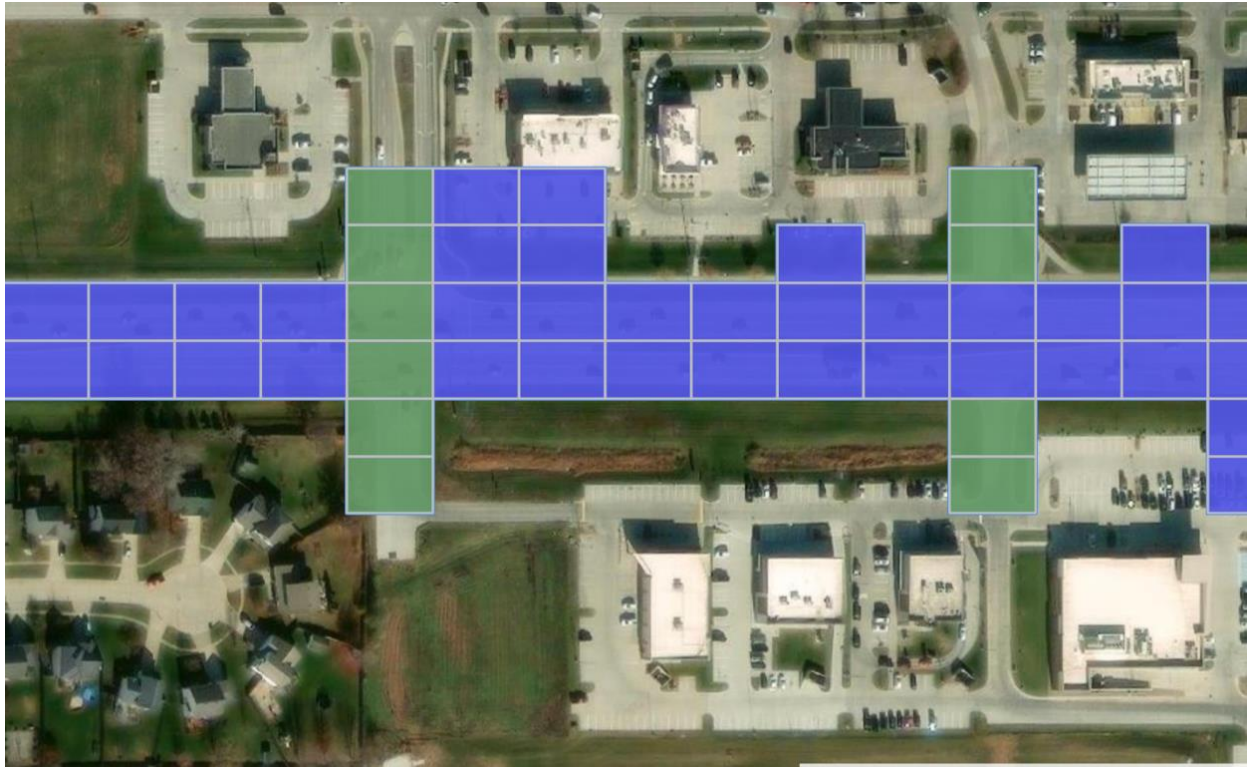

**Figure 8 Image of classficied roadway segments. Straight roadway segments (Blue colored grid boxes), Intersections (Green colored gird boxes)**

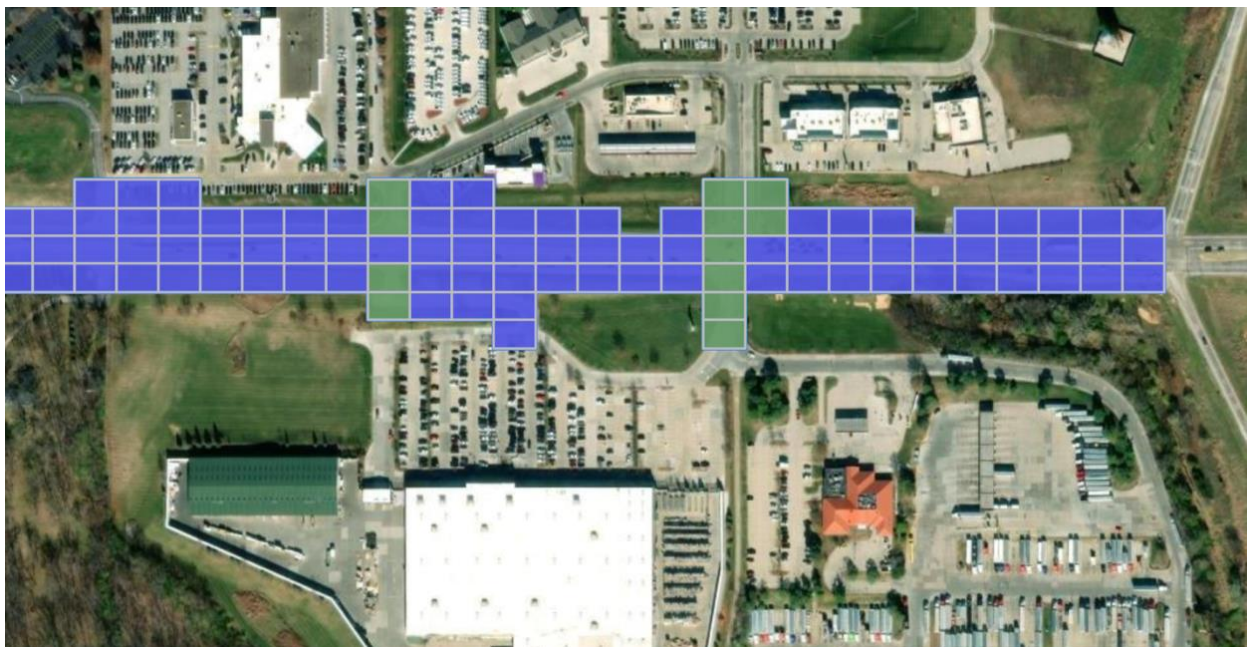

**Figure 9 Image of classficied roadway segments. Straight roadway segments (Blue colored grid boxes), Intersections (Green colored gird boxes)**

**DISCUSSION**

The methodology developed was found to accurately identify intersections and straight road segments. The high precision and recall values indicate a balanced trade-off between correctly identifying instances of each class and minimizing false positives and false negatives. The overall accuracy of 0.86 without colored images and 0.95 with colored images suggests that the model exhibits a strong capability to classify road infrastructure accurately.

The incorporation of waypoint speed information within geohashes, for retraining the model improved in precision, recall, and F1-scores for both the "Intersection" and "Straight" classes. The enhanced accuracy achieved with the new dataset indicates the value of leveraging additional contextual information to improve classification performance.

However, there were instances of misclassifications as shown in **Figure 8**, particularly in cases where the trajectories had irregular shapes or when partially captured within the images. These challenging cases highlight the need for further refinement in the model's ability to accurately classify complex road scenarios.

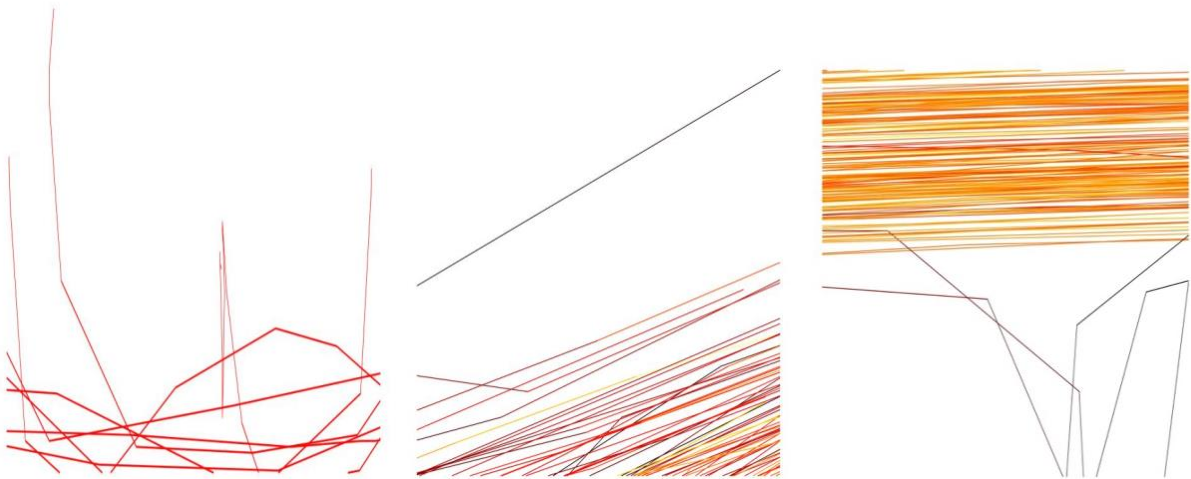

**Figure 8 Sample of Misclassified Images, irregular shaped trajectories (left), partially captured trajectories (middle), misclassified intersection (right).**

**Implications and Future Work**

The proposed methodology presents a novel approach to extract road infrastructure information from connected vehicle trajectory data, combining geohashing and image classification techniques.

The findings suggest potential avenues for further improvement and research. For instance, exploring additional contextual information beyond average speed, such as road features, traffic conditions, and weather data, could enhance the model's accuracy and robustness. Additionally, refining the model to better handle complex road scenarios and irregular trajectories would be impactful.

Overall, the results and discussions presented in this paper contribute to the advancement of road infrastructure extraction methods and offer valuable insights for researchers and practitioners in the field of transportation and connected vehicles. Further research and experimentation will be critical to validate the methodology's effectiveness and its broader application in real-world scenarios.

**CONCLUSIONS**

In this paper, we presented a comprehensive approach to road infrastructure classification using the YOLOv5 model trained on trajectory images generated from geohashes. Our goal was to accurately identify "Intersection" and "Straight" road segments.

After 100 training epochs, the model showed reasonable accuracy, with precision, recall, and F1-scores of 0.73, 0.69, and 0.71 for the "Intersection" class, and 0.90, 0.92, and 0.91 for the "Straight" class, respectively. The overall accuracy achieved was 0.86, providing a strong foundation and opportunities for further refinements.

To enhance the model's performance, we retrained it on a new dataset of colored images, where trajectories within each geohash were color-coded based on waypoint speed. The new approach yielded significant improvements, with precision, recall, and F1-scores reaching 0.98, 0.84, and 0.90 for the "Intersection" class, and 0.94, 0.99, and 0.97 for the "Straight" class, respectively. The overall accuracy achieved by the model trained on the new dataset was 0.95, surpassing the previous accuracy.

The findings highlight the significance of leveraging additional contextual information, such as average speed, to enhance the model's classification performance. The combination of YOLOv5 with colored images based on average speed information within geohashes improved road segment classification accuracy and reliability which improved the identification of true positives and reduced the number of false negatives.

However, certain limitations exist, including the need for further investigation on the generalizability of the model to different datasets and real-world scenarios, especially concerning complex road networks with diverse infrastructure types and irregular vehicle trajectories.

Future research should focus on expanding the dataset to encompass a wider range of road infrastructure instances, exploring the model's performance by considering the integration of other relevant contextual features.

Additionally, extending the model's capabilities to detect various types of intersections, understanding traffic flows at intersections, and addressing challenges with irregular trajectories and partially captured images will further enhance its usefulness in traffic management, urban planning, and transportation infrastructure optimization.

In conclusion, our study demonstrates the potential of a new methodology which relies on connected vehicle trajectory data and a deep learning model to improve the accuracy and ease of classifying roadway infrastructure. These findings contribute to the advancement of road infrastructure classification techniques and open avenues for more sophisticated and effective applications in transportation and urban planning fields. The proposed approach holds promise for advancing road safety network analysis, traffic management, and higher levels of vehicle autonomy.

**REFERENCES**

1. Dovis, Fabio, et al. "Recent advancement on the use of global navigation satellite system-based positioning for intelligent transport systems [guest editorial]." IEEE Intelligent Transportation Systems Magazine 12.3 (2020): 6-9.

2. Kavzoglu, Taskin & Sen, Yunus & Cetin, Mufit. (2009). Mapping urban road infrastructure using remotely sensed images. International Journal of Remote Sensing. 30. 1759-1769. 10.1080/01431160802639582.

3. Hu, Jiuxiang, et al. "Road network extraction and intersection detection from aerial images by tracking road footprints." *IEEE Transactions on Geoscience and Remote Sensing* 45.12 (2007): 4144-4157.

4. Bajcsy, Ruzena, and Mohamad Tavakoli. "Computer recognition of roads from satellite pictures." *IEEE Transactions on Systems, Man, and Cybernetics* 9 (1976): 623-637.

5. Lefler, Nancy, et al. *Model inventory of roadway elements-MIRE, version 1.0*. No. FHWA-SA-10-018. 2010.

6. Gao, Song, et al. "Automatic urban road network extraction from massive GPS trajectories of taxis." *Handbook of Big Geospatial Data*. Cham: Springer International Publishing, 2021. 261-283.

7. Saldivar-Carranza, E. and Bullock, D. (2023) A Data-Driven Intersection Geometry Mapping Technique to Enhance the Scalability of Trajectory-Based Traffic Signal Performance Measures. *Journal of Transportation Technologies*, **13**, 443-464. doi: 10.4236/jtts.2023.133021.

8. Lian, Renbao, et al. "Road extraction methods in high-resolution remote sensing images: A comprehensive review." IEEE Journal of Selected Topics in Applied Earth Observations and Remote Sensing 13 (2020): 5489-5507.

9. Zhang, A., Zhou, W., & Li, J. (2018). Deep learning for remote sensing data: A technical tutorial on the state of the art. IEEE Geoscience and Remote Sensing Magazine, 6(2), 27–60. https://doi.org/10.1109/MGRS.2018.2824958

10. Bakhtiari, Hamid Reza Riahi, Abolfazl Abdollahi, and Hani Rezaeian. "Semi automatic road extraction from digital images." The Egyptian Journal of Remote Sensing and Space Science 20.1 (2017): 117-123.

11. Shi, Wenzhong, Zhaoyuan Yu, and Yuemin Yue. "Evaluation of automatic road extraction from multi-source satellite images in mountainous urban area." International Journal of Remote Sensing 30.2 (2009): 353-372.

12. Zeng, Yuanyuan, Liang Cheng, Manchun Li, Yongxue Liu, and Anshu Zhang. "A survey of image classification methods and techniques for improving classification performance." International Journal of Remote Sensing 33.13 (2012): 4094-4106.

13. Xie, D., Li, F., Yao, B., & Li, G. (2018). Survey of Real-Time Processing Systems for Big Data. In 2018 20th International Conference on High Performance Computing and Communications; IEEE 16th International Conference on Smart City; IEEE 4th International Conference on Data Science and Systems (HPCC/SmartCity/DSS), 1408–1415. https://doi.org/10.1109/HPCC/SmartCity/DSS.2018.00230

14. Gao, Song, et al. "Automatic urban road network extraction from massive GPS trajectories of taxis." Handbook of Big Geospatial Data. Cham: Springer International Publishing, 2021. 261-283.

15. Saldivar-Carranza, E. and Bullock, D. (2023) A Data-Driven Intersection Geometry Mapping Technique to Enhance the Scalability of Trajectory-Based Traffic Signal Performance Measures. Journal of Transportation Technologies, 13, 443-464. doi: 10.4236/jtts.2023.133021.

16. Fathi, Alireza, and John Krumm. "Detecting road intersections from GPS traces." Geographic Information Science: 6th International Conference, GIScience 2010, Zurich, Switzerland, September 14-17, 2010. Proceedings 6. Springer Berlin Heidelberg, 2010.

17. Burkhard, Oliver, et al. "On the requirements on spatial accuracy and sampling rate for transport mode detection in view of a shift to passive signalling data." Transportation Research Part C: Emerging Technologies 114 (2020): 99-117.

18. Bochkovskiy, A., Wang, C.-Y., & Liao, H.-Y. M. (2020). YOLOv4: Optimal Speed and Accuracy of Object Detection. https://arxiv.org/abs/2004.10934

19. Niemeyer, Gustavo. "Geohash." Retrieved June 6 (2008): 2018. *an article about Geohash witnessing and citing G. Niemeyer works, before 2012*

20. Redmon, Joseph, et al. "You only look once: Unified, real-time object detection." Proceedings of the IEEE conference on computer vision and pattern recognition. 2016.

21. https://www.fhwa.dot.gov/policyinformation/hpms/documents/arnold_reference_manual_2014.pdf